\documentclass[conference]{IEEEtran}
\IEEEoverridecommandlockouts
\usepackage{cite}
\usepackage[numbers,sort&compress]{natbib}
\usepackage{amsmath,amssymb,amsfonts}
\usepackage{algorithmic}
\usepackage{amsthm}
\usepackage{enumerate}
\usepackage{amsfonts}
\usepackage{color}
\usepackage{hyperref}
\usepackage[justification=centering]{caption}
\usepackage[ruled,linesnumbered]{algorithm2e}
\usepackage{graphicx}
\usepackage{textcomp}
\usepackage{xcolor}
\usepackage{multicol}
\usepackage{amsthm}
\usepackage{caption}
\usepackage{subfigure}
\usepackage{tabularx}
\usepackage{cite}
\usepackage{url}
\usepackage{setspace}
\usepackage[utf8]{inputenc}
\usepackage{multirow}
\usepackage{tikz}
\usetikzlibrary{positioning}

\usepackage{booktabs}

\def\BibTeX{{\rm B\kern-.05em{\sc i\kern-.025em b}\kern-.08em
    T\kern-.1667em\lower.7ex\hbox{E}\kern-.125emX}}
\begin{document}


\title{Streaming Traffic Flow Prediction Based on Continuous Reinforcement Learning }

\author{\IEEEauthorblockN{Yanan Xiao$^{1}$, Minyu Liu$^{1}$, Zichen Zhang$^{1}$, Lu Jiang$^{1,*}$, Minghao Yin$^{1}$, Jianan Wang$^{2,*}$,}
\IEEEauthorblockA{\textit{$^1$Information Science and Technology, Northeast Normal University, Changchun} \\
\textit{$^2$College of Physics, Northeast Normal University, Changchun} \\
\textit{\{xiaoyn117, liumy333, zhangzc482, jiangl761, ymh, wangjn\}@nenu.edu.cn }}
\textit{Corresponding author*}
}

\maketitle


\begin{abstract}
Traffic flow prediction is an important part of smart transportation. The goal is to predict future traffic conditions based on historical data recorded by sensors and the traffic network. As the city continues to build, parts of the transportation network will be added or modified. How to accurately predict expanding and evolving long-term streaming networks is of great significance. To this end, we propose a new simulation-based criterion that considers teaching autonomous agents to mimic sensor patterns, planning their next visit based on the sensor's profile (e.g., traffic, speed, occupancy). The data recorded by the sensor is most accurate when the agent can perfectly simulate the sensor's activity pattern. We propose to formulate the problem as a continuous reinforcement learning task, where the agent is the next flow value predictor, the action is the next time-series flow value in the sensor, and the environment state is a dynamically fused representation of the sensor and transportation network. Actions taken by the agent change the environment, which in turn forces the agent's mode to update, while the agent further explores changes in the dynamic traffic network, which helps the agent predict its next visit more accurately. Therefore, we develop a strategy in which sensors and traffic networks update each other and incorporate temporal context to quantify state representations evolving over time. 
Along these lines, we propose streaming traffic flow prediction based on continuous reinforcement learning model (ST-CRL), a kind of predictive model based on reinforcement learning and continuous learning, and an analytical algorithm based on KL divergence that cleverly incorporates long-term novel patterns into model induction. Second, we introduce a prioritized experience replay strategy to consolidate and aggregate previously learned core knowledge into the model. The proposed model is able to continuously learn and predict as the traffic flow network expands and evolves over time. Extensive experiments show that the algorithm has great potential in predicting long-term streaming media networks, while achieving data privacy protection to a certain extent.
\end{abstract}
\section{Introduction}
A long-term streaming network refers to the fact that the topology of the transportation network is continuously added and removed over a long period of time (ie, years and months), and nodes (sensors) are constantly changing, in line with real-world scenarios. In this paper, we investigate the integration of continuous reinforcement learning in long-term traffic flow network analysis. Classical traffic flow prediction is based on historical traffic data flow and static traffic network topology to predict future traffic flow conditions. In real-world problems, traffic networks and traffic flows are constantly changing. Traffic sensors deployed in transportation facilities record the expansion of the transportation network and the evolution of traffic flow. How to model and predict complex spatial dependencies and dynamic trends of spatio-temporal patterns is a pressing problem: ST-CRL, which aims to generate modeling and update, and finally make predictions.

Existing contributions to traffic flow prediction include CNNs, LSTMs, Transformers \cite{zhang2017deep,li2017diffusion,wang2022lifelong}, and various cutting-edge methods have been introduced into the field to improve the accuracy of short-term prediction in static networks. Part of the research focuses on modeling the communication network using GNN for traffic flow prediction, and has achieved great success by mining space-time correlation \cite{yu2017spatio}. For example, VAR, SVR \cite{zivot2006vector,castro2009online} reduce the prediction to a single time series forecast. The spatio-temporal correlation between different locations has failed to be captured using statistical modeling approaches. In addition, training such models for all locations leads to poor performance, while using models for each location increases consumption time and storage costs. With the development of deep learning, researchers have applied various frameworks to overcome the difficulties in understanding spatio-temporal data patterns. Part of the research proposes to use reinforcement learning learning agents to learn autonomously to enrich spatial semantics to simulate streaming spatio-temporal data \cite{2020Incremental,DBLP:conf/kdd/WangFZWZA18,DBLP:journals/tist/WangFZLL18,DBLP:conf/kdd/WangFXL19}. There are also some studies that propose feasible algorithms that can be used for specific communication prediction problems \cite{wang2020sccwalk,pan2023improved}. Although these methods have achieved impressive results on the traffic prediction task, most of the research cannot be directly applied to the prediction of long-term streaming networks.

Continuous learning, also known as incremental learning and lifelong learning, is roughly equivalent in concept and has the ability to continuously process a continuous flow of information in the real world, and retain or even integrate and optimize old knowledge while absorbing new knowledge. Existing methods can be roughly divided into three categories: regularization-based methods \cite{kirkpatrick2018reply}, replay-based methods \cite{scheller2020sample}, and parameter isolation methods \cite{rusu2016progressive}. In recent years, continuous learning has been applied in the field of dynamic graph learning, and continual GNN and ER-GNN have achieved good results in incremental learning that only considers the GNN model when the graph structure changes. However, when the temporal patterns of nodes and network topology evolve simultaneously, they cannot handle efficient updating of spatio-temporal models in transportation networks.

To capture the expansion and evolution of long-term streaming networks, the simplest approach is to retrain the model as the road network structure is updated, a continuous learning called incremental upper bound. However, the consumption of resources is huge and the newly trained model will lose historical information, and the relative increase of the parameters of the new model poses serious challenges to efficiency and performance. Transferring existing network knowledge to the new model for resource saving and data protection is an intuitive architecture, given that the old model already produces good predictions. There are two issues to consider when using the old model directly. One is how to model new data. As streaming networks expand and evolve, large discrepancies between old and new data can reduce the accuracy of predictions, or even lead to catastrophic forgetting. The topology of the transportation network is also different from before. Therefore, three challenges need to be faced: how to model long-term streaming networks, how to extract useful knowledge from old models and fuse them, and how to efficiently capture new data models.

After a comprehensive analysis, we identified a better criterion based on imitation: teaching autonomous agents to imitate sensors based on individual data (e.g., traffic, speed, occupancy) recorded by the sensors. Based on the sensor records, the traffic values for the next moment are planned. Traffic flow prediction is most accurate when the agent can perfectly replicate the sensor activity pattern. Emerging reinforcement learning can train agents to plan their next actions to function in their environment. The state of the environment is a fused representation of sensor and traffic network dynamics, and this capability offers great potential for implementing imitation-based criteria for more accurate traffic flow prediction.

We therefore formulate the problem as a continuous reinforcement learning framework, where the agent is the next moment flow value predictor, the action is the next set of flow values in the sensor, and the environmental state is the fused representation of the sensor patterns. For example, traffic, speed, occupancy) and traffic network dynamics. The action taken by the agent to predict the flow value changes the environment, causing the sensors to record data updates, while the agent further explores changes in the dynamic traffic network, which helps the agent to more accurately predict the traffic flow at the next moment. The action pays off by reducing the gap between proxy activity patterns and actual traffic. Re-enacted, our new goal is to use a continuous reinforcement learning framework to extract traffic network representations and traffic flow predictions in environmental states by continuous reinforcement learning from long-term streaming networks.

To improve efficiency and effectiveness, based on the above architecture, we have further designed the model by integrating unknown patterns of new nodes and changing patterns of old nodes. To discover the differences between the old and new models, we propose an algorithm based on KL scatter to measure the changes of node characteristics with the model. To consolidate the traffic knowledge of the old nodes and transfer it to the new model for better prediction, we employ a prioritized experience replay strategy from a data utilization perspective.By combining the two approaches in the model, a balance is struck between learning new traffic patterns from new data and maintaining historical knowledge learned from previous data.

Our main contributions are summarized as follows. (1) We propose a new imitation-based criterion for traffic flow prediction: the more the agent can imitate the standard sensor model, the more accurate the traffic prediction will be. (2) Motivated by model-based criteria, we reshape the long-term streaming network prediction problem into a continuous reinforcement learning framework, where the agent is the next traffic value predictor and the action is the next sequence of traffic values in the sensor, The state of the environment is a dynamically fused representation of the sensor and traffic network. The reward for an action is the agent's ability to predict the accuracy of traffic. (3) We demonstrate extensive experimental results on real traffic networks to demonstrate performance improvements. (4) Our framework can be generalized to learning tasks with long-term network structure, including but not limited to speed, traffic, congestion, etc., to support model building of intelligent transportation.
\section{Background and Problem Definition}
We first introduce the key definitions and the problem statement.Then, we show the overview of the proposed framework, followed by the discussion of difference with literature. All the notations are summarized in Table 1.

\subsection{Key Components of Reinforcement Learning}
In our problem setting, we show that predicting traffic flow can be formulated as a continuous reinforcement learning framework as follows:
\begin{enumerate}

\item {\bf Agents.} We treat the next traffic value predictor as a proxy, i.e. predicting the traffic flow in the temporal context. Composition of all sensors based on the current state of the environment.
\item {\bf Actions.} 
The action space $A$ is the prediction range of the agent for the traffic flow. The action $a_t \in A$ represents our prediction of traffic flow in the simulated time context $t$. To further simplify the action space, set the traffic range threshold to 5 categories, i.e. [0, 1, 2, 3, 4] to represent the agent's prediction of traffic in different states.
\item {\bf Environment.} The environmental state is a fused representation of sensor patterns and traffic network dynamics in long-term streaming networks. The action taken by the agent at time $t$ (i.e. traffic flow prediction) changes the environment, resulting in an update of the agent's mode, while the agent further explores changes in the dynamic traffic network, which helps the agent to more accurately predict the next visit. Specifically expressed as $G=(G_1, G_2,...,G_t)$, where $G_t=G_{t-1}+\Delta$, $\Delta$ represents the expansion or deletion of traffic network nodes in the temporal context.
\item {\bf State.} The state $s$ represents the mode representation of the current agent (i.e. sensor) under the traffic network at time $t$, specifically the traffic, speed and occupancy rate on the sensor at the time of deployment $t$, as well as the current local connected network information.
\begin{figure*}[!t]
	\centering
	\includegraphics[width=1\linewidth]{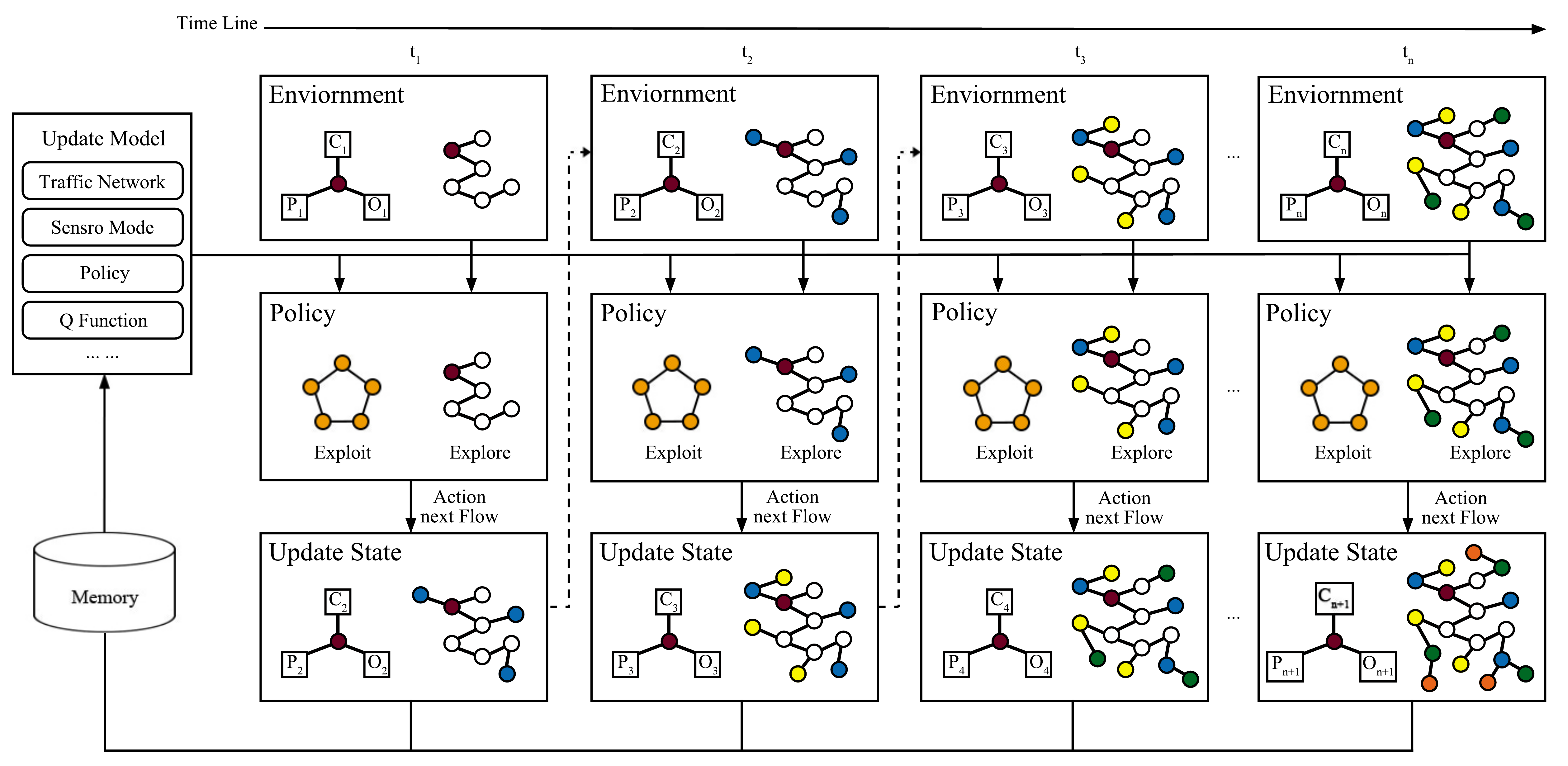}
	\captionsetup{justification=centering}
	\caption{Framework Overview.}
	\label{fig:framework overview}
\end{figure*}
\item {\bf Reward.} The agent predicts different traffic flow networks through the specific flow, speed and occupancy rate under the state $s$, we define the reward $r$ as (i) $r_p$, the difference between the actual flow value and the predicted flow value ; (ii) $r_c$, the speed, the traffic flow is proportional to the speed; (iii) $r_o$, the occupancy rate, the traffic flow is inversely proportional to the occupancy rate, here is the reciprocal. The reward can be denoted as:
\begin{equation}
    r=\lambda_p \times r_p +\lambda_c \times r_c + \lambda_o \times r_o 
\end{equation}
where $\lambda_p$, $\lambda_c$ and $\lambda_o$ denote the weights of the following $r_p$, $r_c$ and $r_o$ terms. We use the class difference between predicted and actual traffic flow as reward $r_p$.
\end{enumerate}
\subsection {Problem Statement}
In this paper, we study the long-term streaming network prediction problem. Due to the expanding and evolving nature of long-term traffic networks and traffic flows, we reformulate the traffic flow prediction problem as a continuous reinforcement learning problem that combines the data utilization of continuous learning and the modeling representation of reinforcement learning. Formally, given the state of the environment (traffic network and traffic flow pattern) before time t as input, the goal is to predict the traffic flow after the topology change of the road network at time $t_1$ as output. Sensor patterns are updated incrementally with the environment to provide accurate predictions.

\subsection {Framework Overview}
Figure 1 shows an overview of our proposed framework, which includes the following key components: (1) Environment state updates. We consider agent patterns to interact with the environment at a given time context. Specifically, at each step, the agent state is updated by making a traffic flow prediction with environmental influence at the given time context; instead, the environment forces the agent mode to update at the given time context. (2) Strategy learning. In order to discover the difference between the old and new models, we propose an algorithm based on KL divergence to measure the change of node features with the model. In order to consolidate the traffic knowledge of the old nodes and transfer it to the new model for better prediction, we adopt a prioritized experience playback strategy from the perspective of data utilization. Predictions are evaluated against the gap between actual and predicted traffic flow, which in turn forces the next visit planner (agent) to imitate sensor preferences and patterns to satisfy imitation-based learning criteria. (3) Feedback. The agent accepts the environment and gives feedback, i.e. predicts traffic flow. Feedback is represented by a reward $r$. (4) Policy update. After each time step, we have the current state $s_t$, the reward feedback from the environment $r_t$, and the next environment state $s_{t+1}$ as a tuple $(s_t, a_t, r_t, s_{ t +1})$ storage and sampling. (5) Repeat steps (1)-(4).

\subsection {Comparison with literature}
Despite the impressive results in traffic flow prediction, the main problem is that most methods aim to predict short-term static network structure without long-term self-update capability. The prediction of long-term streaming media network is of great significance to the construction of smart transportation, which can save a lot of human resources and costs. Therefore, in our work, continuous reinforcement learning is used for traffic flow network modeling and achieves good communication prediction performance through a continuous decision-making process. Unlike static transportation networks, long-term streaming networks are constantly expanding and evolving. Therefore, we provide rich semantics based on time-step, integrated sensor recordings of traffic networks and traffic flows to better understand agent patterns and communication prediction preferences.
\section{Method}
This section details our empirical evaluation of the proposed method
on real-world data.
\subsection{Network Structure}Dueling DQN \cite{mnih2013playing} is widely exploited to learn policies by leveraging deep neural networks. In order to get the exact value of reward $r$, take state $s$ as input, we use the $Q-value$ function to represent the expected return after selecting the action (traffic flow category) in the current state according to the policy $\pi$.

\textbf{Policy Design}. Actions in state st are represented by $Q-values$. Maximize cumulative reward by Bellman equation:
\begin{equation}
Q^{\pi}(s_t, a_t) = \mathbb{E}[r_t +\gamma \mathbb{E}[Q^{\pi}(s_{t+1}, a_{t+1})]
\end{equation}
where $\gamma$ is the discount factor. The $Q-value$ uses the temporal difference(TD)-based method recursively:
\begin{equation}
    \begin{aligned}
        & Q^{\pi}(s_t, a_t) =  Q^{\pi}(s_t, a_t) + \\
        & \alpha[r_t + \gamma max_{a_{t+1}}Q^{\pi}(s_{t+1}, a_{t+1})-Q^{\pi}(s_t, a_t))]
    \end{aligned}
\end{equation}
Where $\alpha$ represents the learning rate.

These elements form the experience tuple at time step $t$, $(s_t, a_t, r_t, s_{t+1})$. The agent draws a batch of memory from the experience replay buffer and computes the following loss function:
\begin{equation}
    Loss(\theta) = E_{\mathcal{B}}[y- Q(s_t, a_t; \theta))^2]
\end{equation}
\begin{equation}
    y = r_t + \gamma max_{a_{t+1}}Q^{\pi}(s_{t+1}, a_{t+1})
\end{equation}
where and $\theta$ is a parameter of the common part of the neural network structure, and $y$ definite as above.

As shown in Figure 2, we link sensor patterns and local link matrix information as the input of a fully connected (FC) layer, then, FC will map a given state $s$ to output $Q(s, a)$, policy selection $Q (s,a)$ the highest value, and use its $a$ as the prediction result.

\subsection{Prioritized Experience Replay.} The training strategy consists of two stages: (1) priority assignment, assigning each data  $(s_t, a_t, r_t , s_{t +1})$ sample a priority score $p_r$; (2) sampling strategy, select data samples from replay buffer for training. The lower the reward obtained by the agent, the better the next-visit planner (agent) mimic the sensor mode, and the smaller the contribution of data samples to policy training. I Therefore, formally, for each data sample $(s_t, a_t, r_t , s_{t +1})$, the reward-based priority score $p_r$ is defined as:
\begin{equation}
    p_r(s_t, a_t, r_t , s_{t +1})=r
\end{equation}

\subsection{Important Sampling.} After we get the priority score, we can't simply select the largest sample to update, because this will lead to too small samples will never be selected, and it is easy to quickly over fit, we use the exponential form to adjust the probability distribution of sampling.After each sample $i$ has an important metric $p_i$, adjust the probability distribution of sampling in the form of real numbers:
\begin{equation}
    P(i)=\frac{p_i}{\sum_k p_k}
\end{equation}
where $P(i)$ is the sampling probability of the $k$-th data sample given the corresponding priority score. We select the top 5\% nodes with the highest scores as candidates for learning new patterns. These nodes have perfectly simulated agent (sensor) patterns and made accurate predictions.

In the traffic flow prediction problem, another problem arises when the time step changes, the model may forget what it has learned previously, because it only focuses on new patterns. Catastrophic forgetting prompts the model to consolidate previous knowledge, and old nodes in the transportation network should be replayed. A simple strategy is to randomly sample traffic flow samples from old nodes, but since these samples are highly randomized, the error is large. One of the existing methods is to trace the source of this knowledge, that is, historical traffic data. The second is to force the current training model to be isolated from the old model parameters. We have recorded samples of perfectly simulated sensor patterns when employing the above-mentioned prioritized experience replay and importance sampling strategies. Motivated by the first approach, such samples are used for the review of historical information.
\begin{figure}[!t]
	\centering
	\includegraphics[width=1\linewidth]{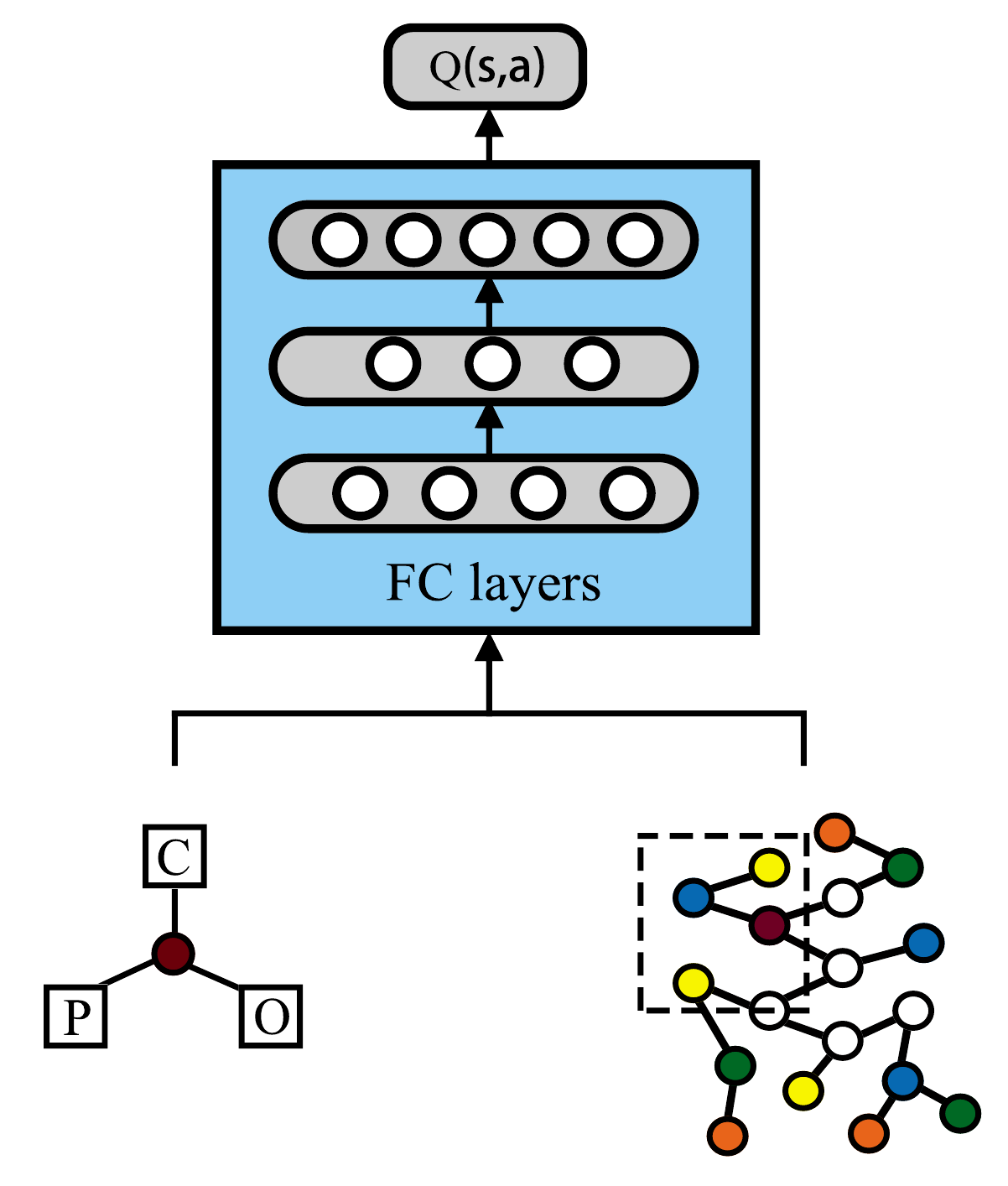}
    \vspace{-0.25cm}
	\captionsetup{justification=centering}
	\caption{Network Structure.}
    \vspace{-0.25cm}
	\label{fig:q-netwrok}
\end{figure}
\subsection{Evolution Pattern Detection}On the one hand, transportation networks expand over time, i.e. sensor connectivity changes over time. So traffic between old and new nodes (sensors) can vary greatly. We design a detection algorithm to detect these nodes. Using $P_{G_t}$ and $P_{G_{t-1}}$ to describe the distribution of the eigenvalues of two nodes, the purpose is to analyze the similarity of the previous two distributions and reveal the change pattern of the old and new nodes. Therefore, KL divergence is introduced as a similarity measure:
\begin{equation}
    KL(G_t||G_{t-1})=\mathbb{E}\left [log \left [\frac{P_{G_t}}{P_{G_{t-1}}} \right ]\right ]
\end{equation}
If a node's KL scores is very high, its pattern can be considered to have evolved substantially. Finally, we select the top 10\% nodes with the highest KL scores as candidates for learning new patterns.

\begin{figure*}[!t]
\centering  
\subfigure[MAE]{
\label{Fig.sub.1}
\includegraphics[width=4.3cm,height = 2.8cm]{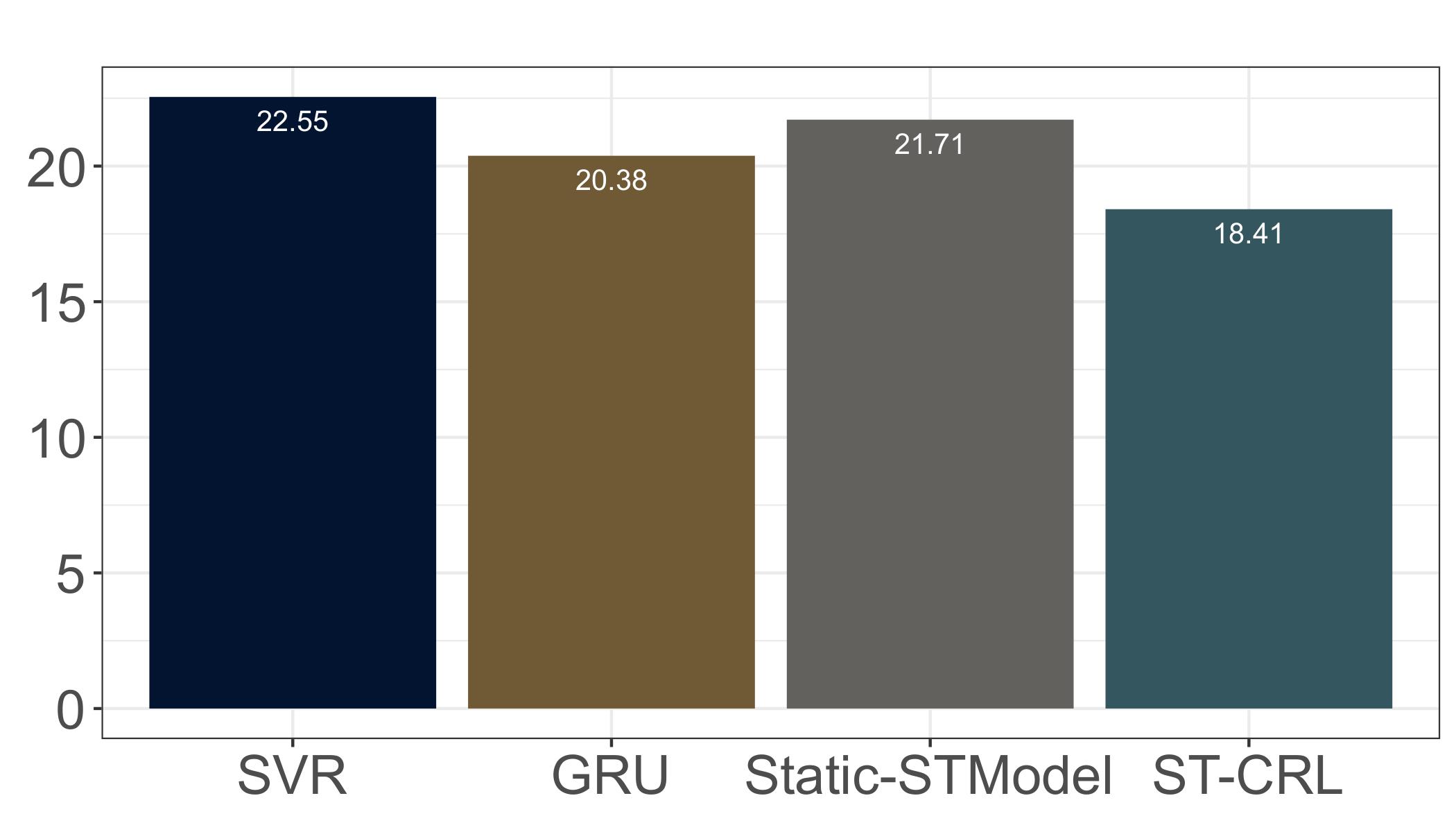}}\subfigure[RMSE]{
\label{Fig.sub.2}
\includegraphics[width=4.3cm,height = 2.8cm]{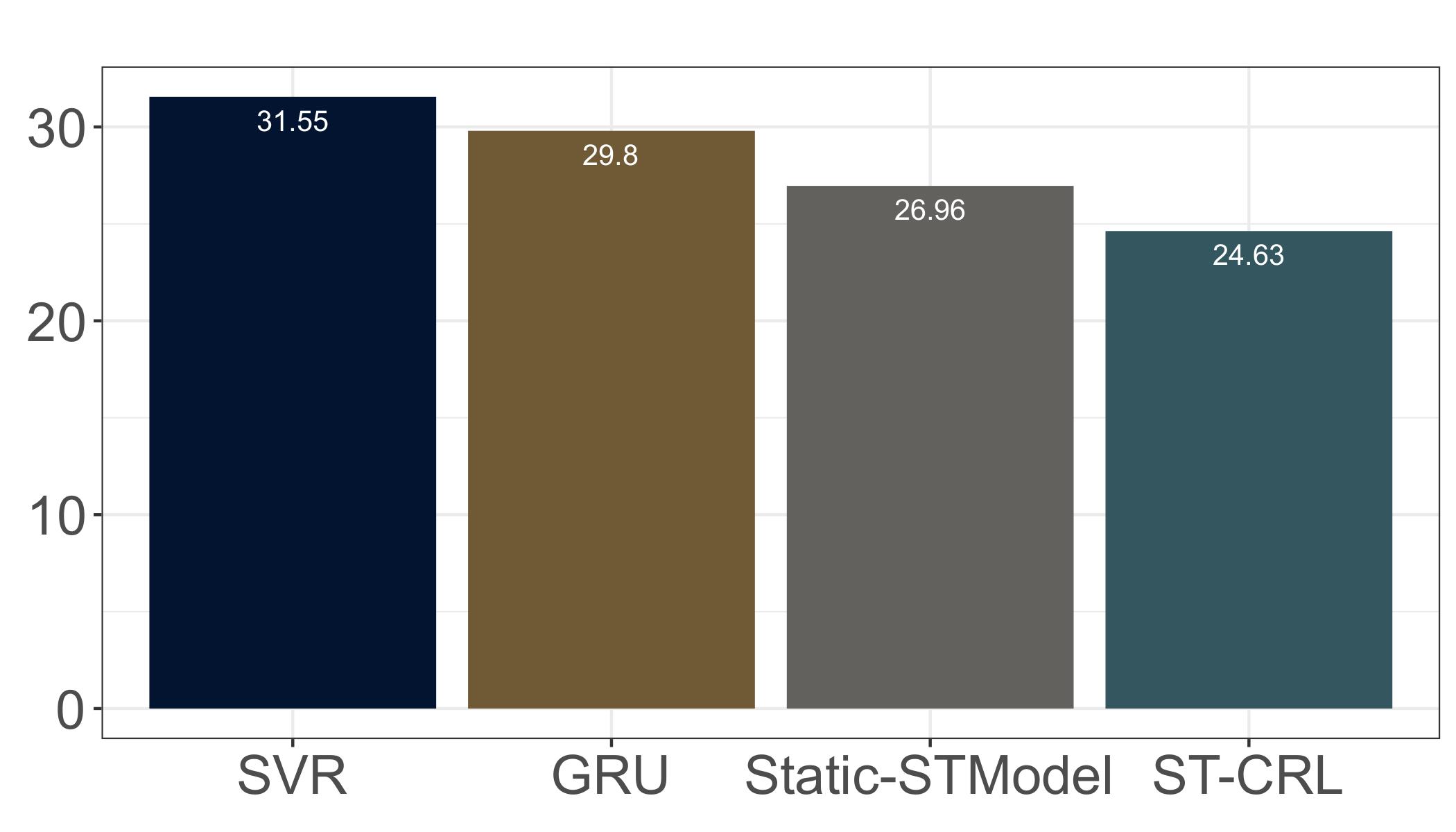}}\subfigure[MAPE]{
\label{Fig.sub.2}
\includegraphics[width=4.3cm,height = 2.8cm]{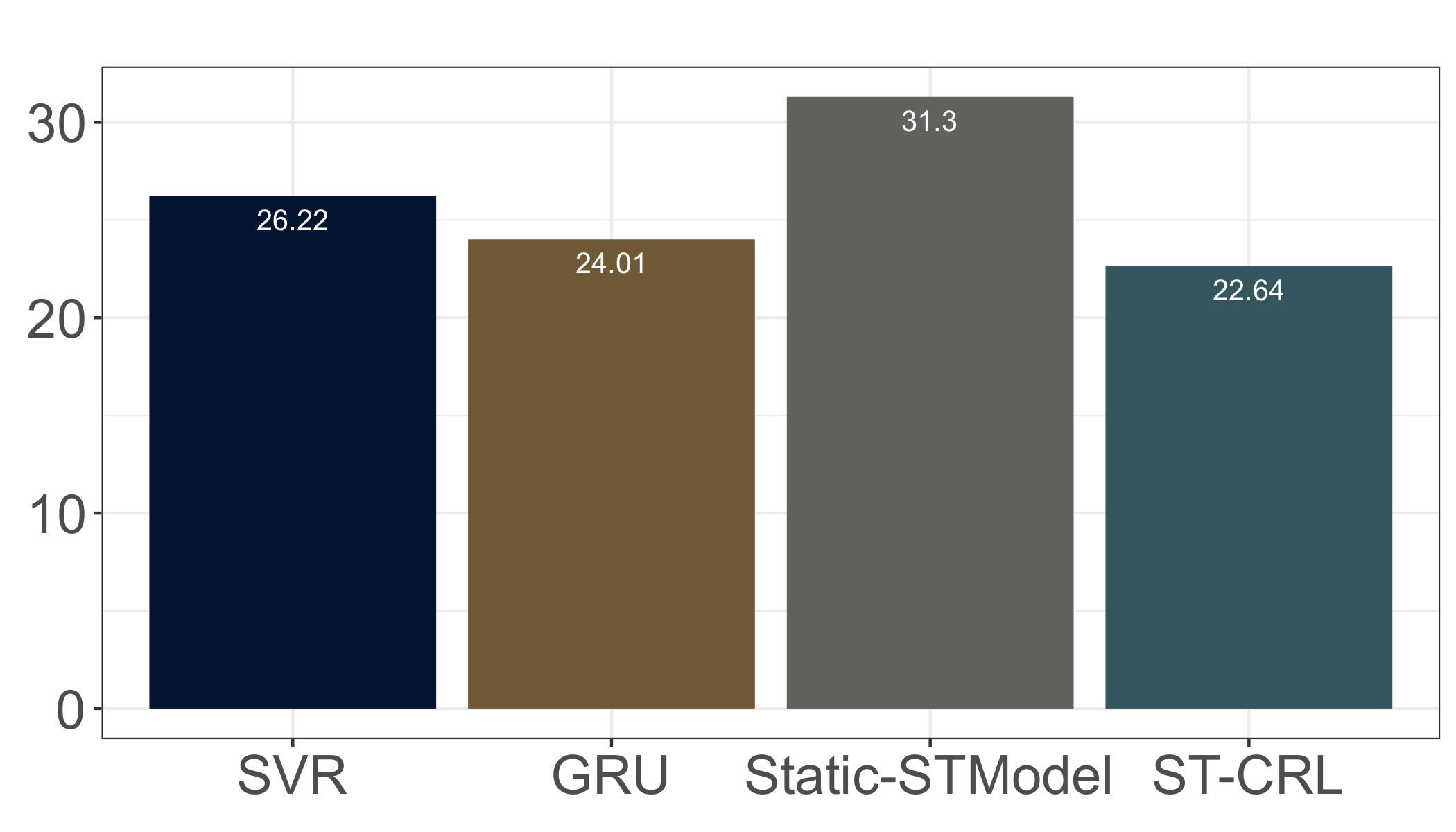}}\subfigure[Total Time]{
\label{Fig.sub.2}
\includegraphics[width=4.3cm,height = 2.8cm]{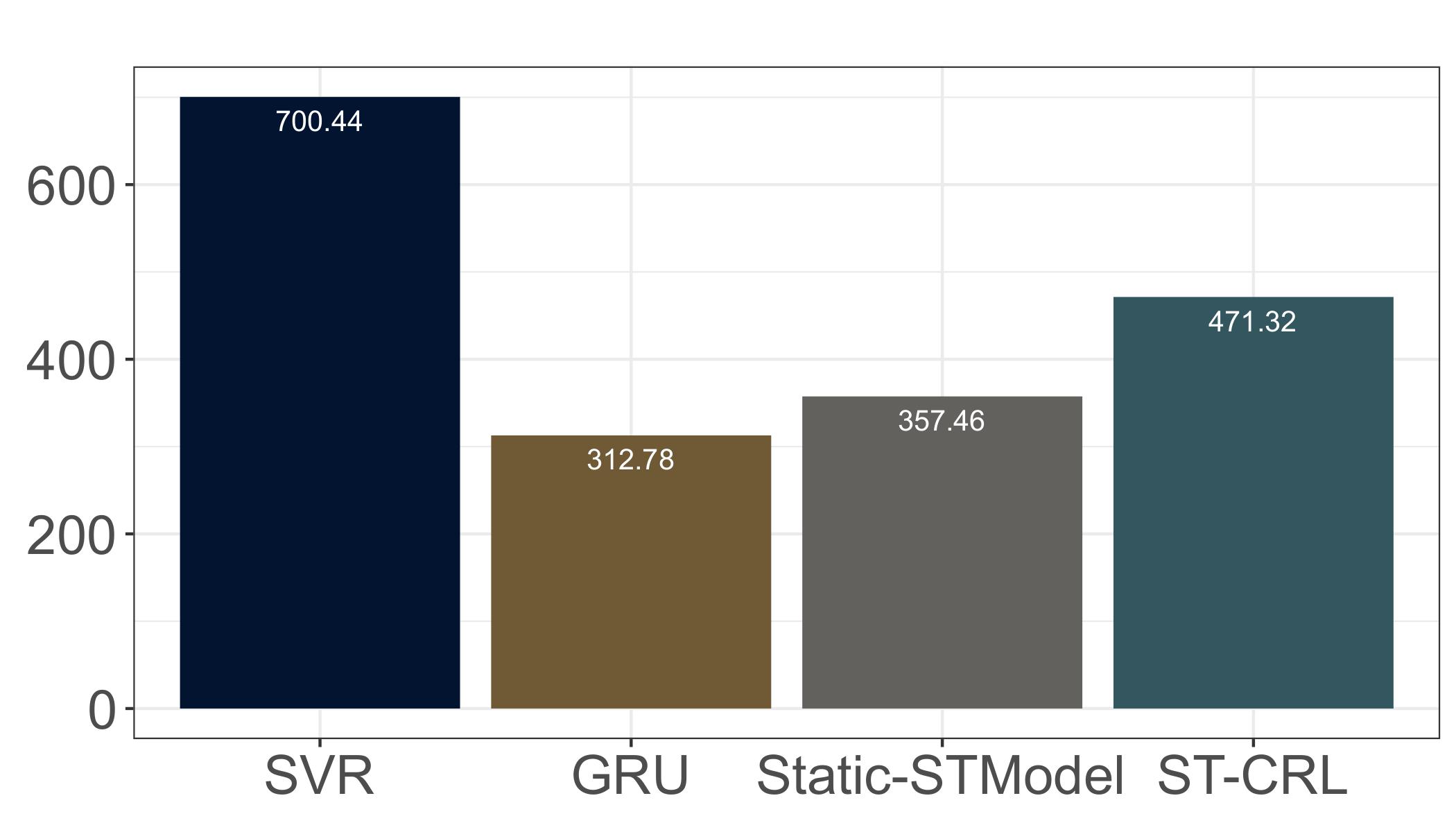}}
\caption{15min on category}
\label{1}
\end{figure*}
\begin{figure*}[!t]
\centering  
\subfigure[MAE]{
\label{Fig.sub.1}
\includegraphics[width=4.3cm,height = 2.8cm]{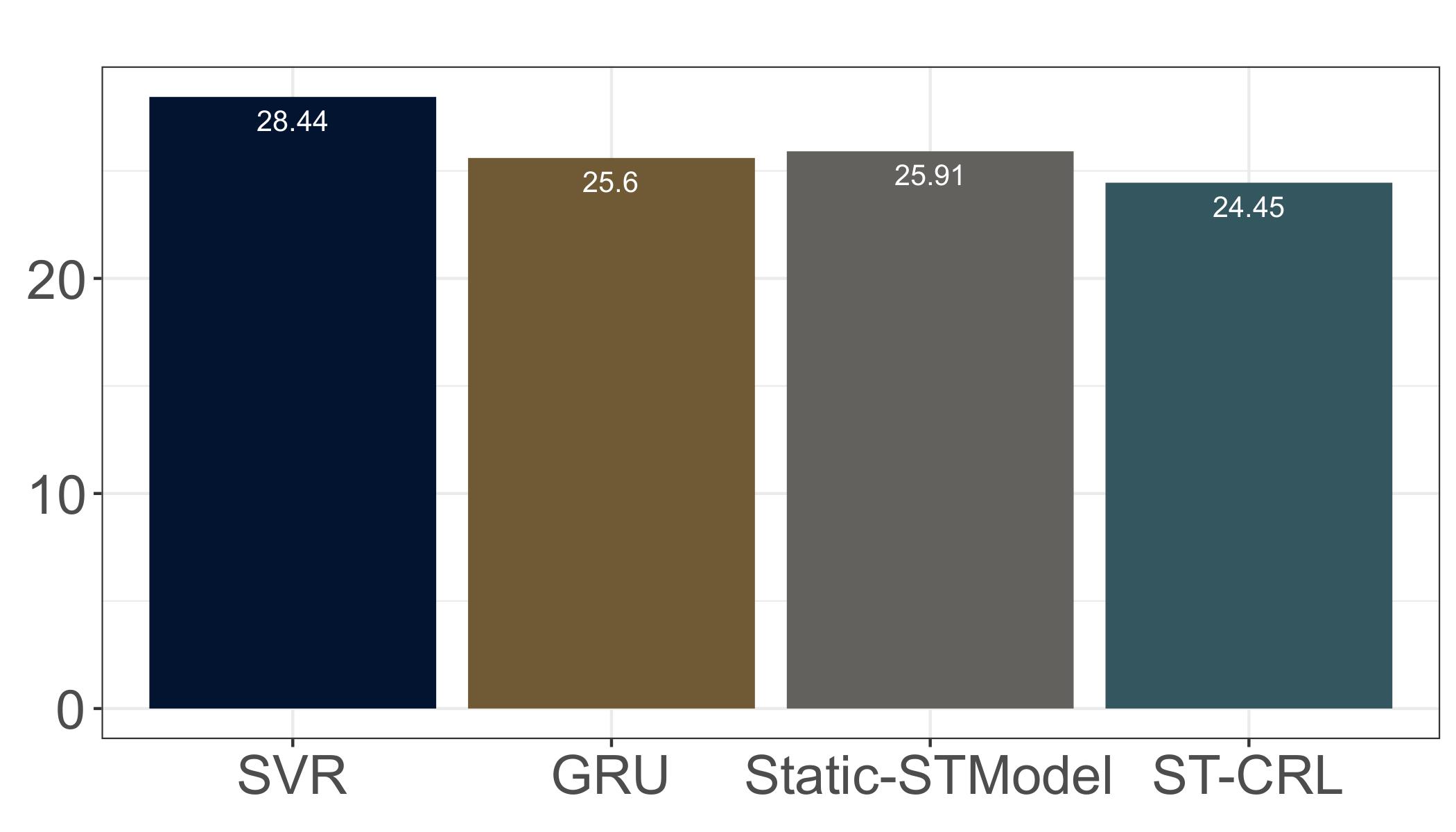}}\subfigure[RMSE]{
\label{Fig.sub.2}
\includegraphics[width=4.3cm,height = 2.8cm]{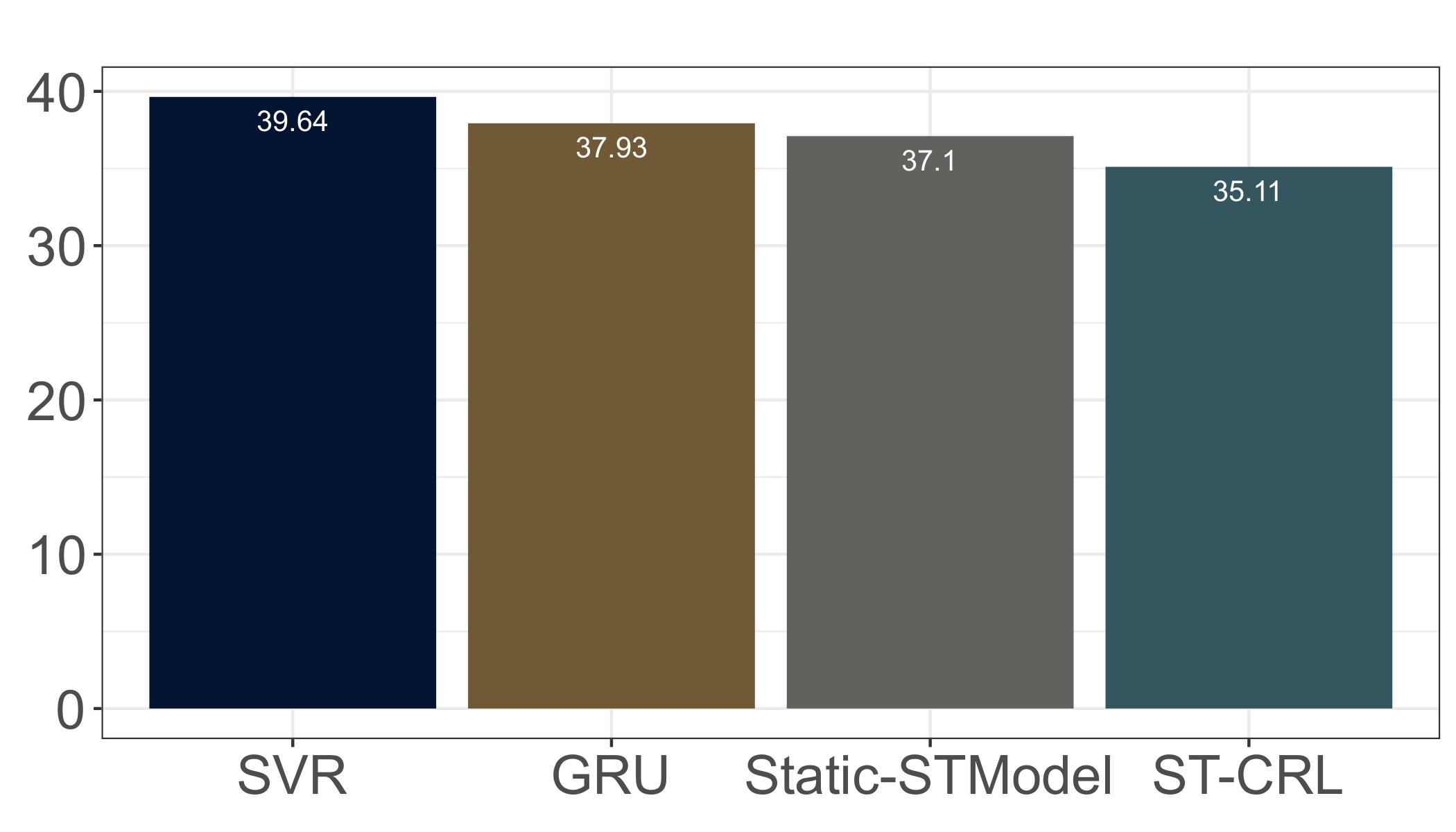}}\subfigure[MAPE]{
\label{Fig.sub.2}
\includegraphics[width=4.3cm,height = 2.8cm]{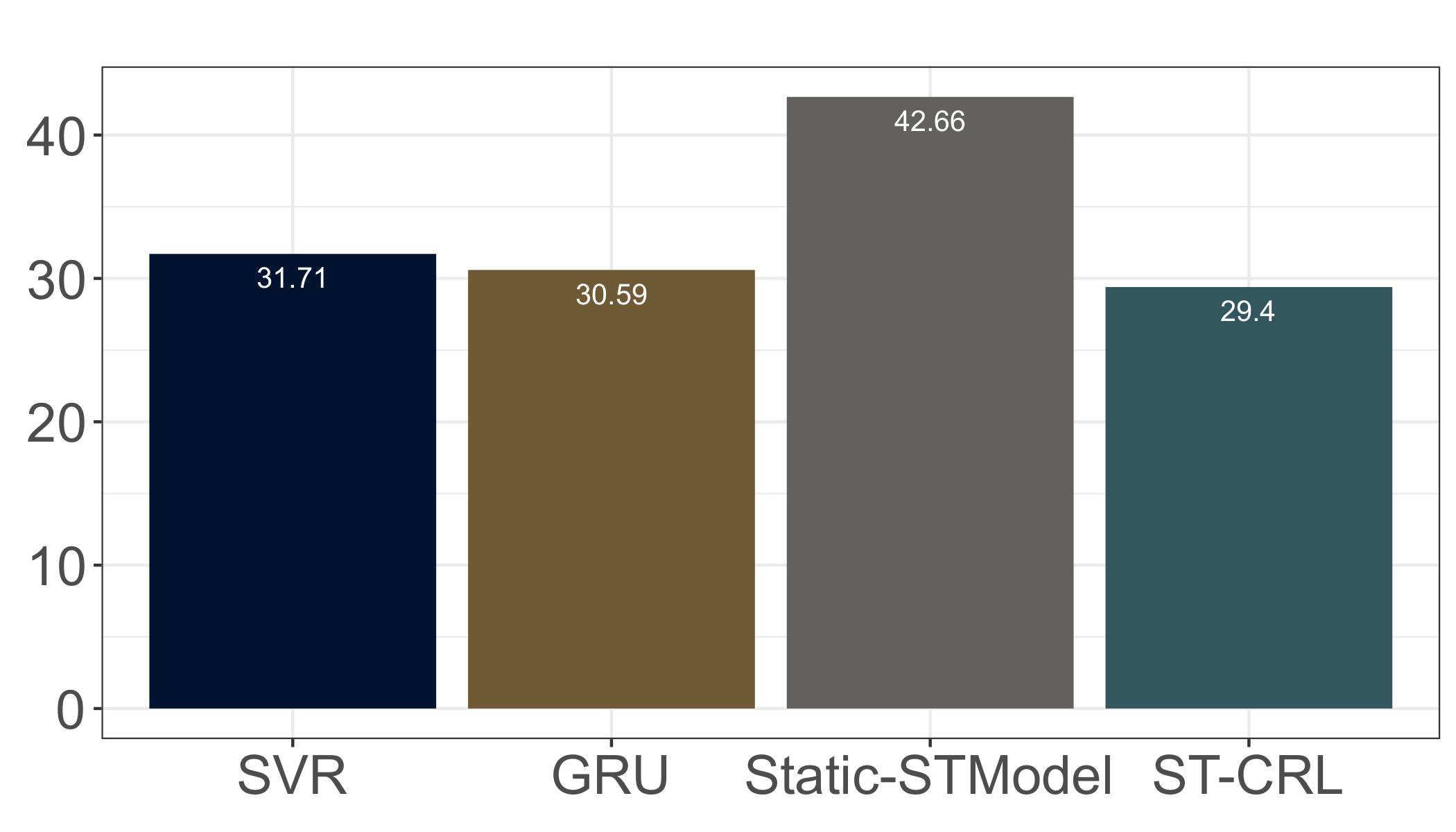}}\subfigure[Total Time]{
\label{Fig.sub.2}
\includegraphics[width=4.3cm,height = 2.8cm]{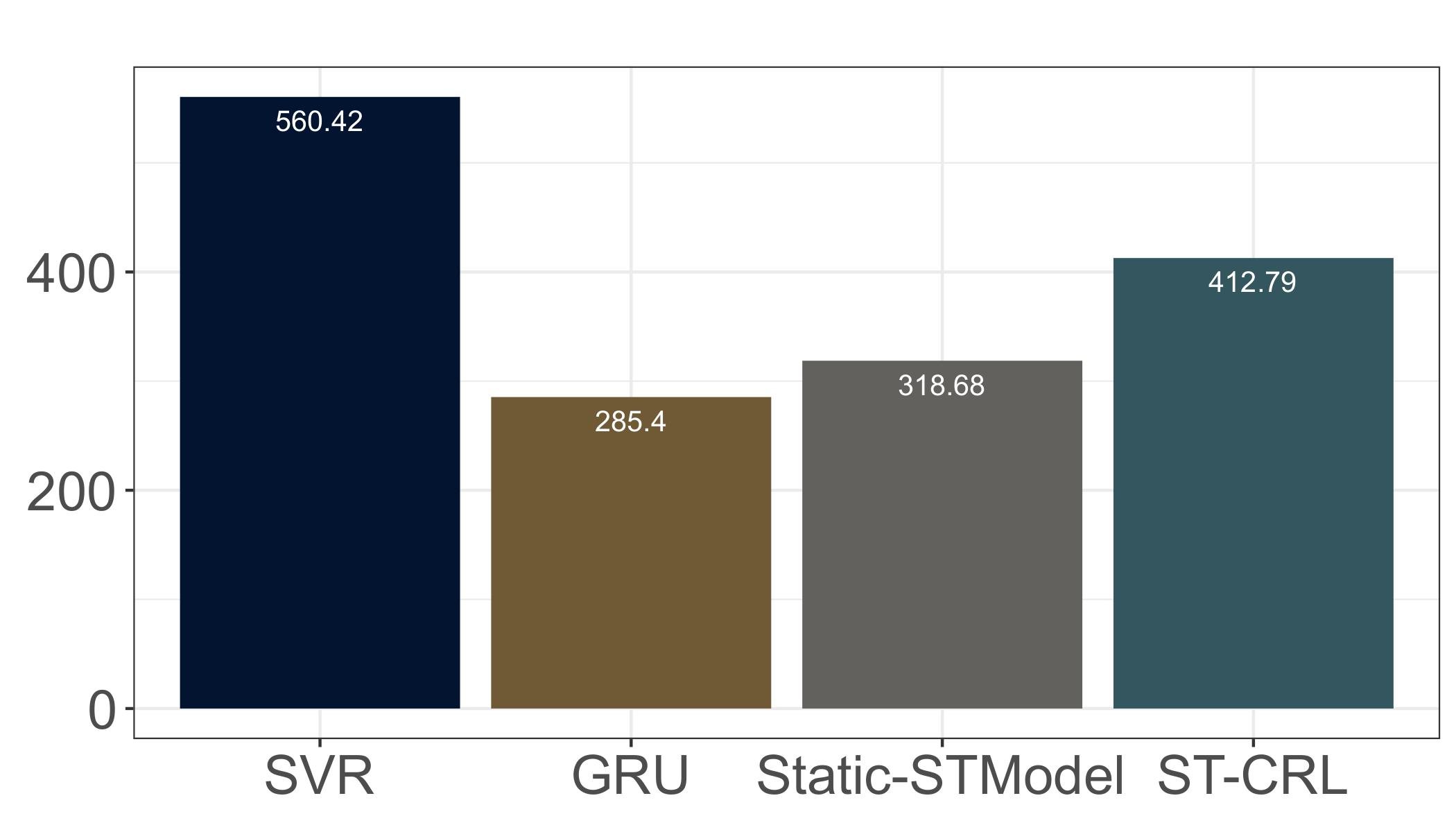}}
\caption{60min on category}
\label{1}
\end{figure*}
\section{Experiment}
In this section, we present an empirical evaluation of the proposed ST-CRL and other baselines on real-world long-term streaming network data.
\subsection{Data Description}We conducted some experiments on the real data set PEMS3 stream \cite{chen2001freeway}, collected by the Highway Performance Measurement System in California, collected data every 30 seconds through the sensor, aggregated the data in 5 minutes, and recorded the PEMS3 stream containing 2011 to 2017 Traffic flow data in the North Central region, while recording flow, occupancy, speed and travel time for each lane (one station typically serves sensors in all lanes at one location). The average time series for 2011-2017 is 15264. These data provide sufficient data and spatio-temporal information. The specific information is shown in Table 1. The annual data is divided into training, validation and test sets in a 6:2:2 ratio in the time dimension. Our task is to predict the traffic flow for the next hour using data from the previous hour. The optimizer learning rate is 0.001, the batch size is 128, and train until convergence (floating around 100 epochs).
\begin{table}[htbp]
	\centering
	\caption {The statistic of PEMSD3-Stream dataset.}
	\scalebox{0.9}{
	\begin{tabular}[t]{c|c c c c c c c}
		\hline
		\textbf{Year} & \textbf{2011} & \textbf{2012} & \textbf{2013} 
		& \textbf{2014} & \textbf{2015} & \textbf{2016} & \textbf{2017} \\ \hline
		Nodes & 655 & 715 & 786  & 822 & 834 & 850 & 871 \\ \hline
		Edges & 1577 & 1929 & 2316 & 2536 & 2594 & 2691 & 2788 \\ \hline
	\end{tabular}}
	\label{data}
\end{table}

\subsection{Evaluation Metrics}To evaluate the performance of our framework, we employ mean absolute error (MAE), root mean square error (RMSE), and mean absolute percentage error (MAPE) as measures of effectiveness. Use total training time and training time for individual nodes to reveal model efficiency.
\subsection{Baseline Methods}
\textbf{SVR \cite{castro2009online}.} Support Vector Regression adapts support vector machine for regression task.

\textbf{GRU \cite{cho2014learning}.} Gated Recurrent Unit is an
RNN model that leverages gated mechanism.

\textbf{Static-STModel \cite{chen2021trafficstream}.}: We utilize data at the first year (i.e.,2011) to train a surrogate model, and forecast traffic flow after 2011 using the surrogate model directly.

\subsection{Overall Comparison}In this section, we compare ST-CRL with other baselines on traffic forecasting tasks and discuss the performance and efficiency results. Figures 2 and 3 show the spatio-temporal (i.e., next 15, 60 minutes) traffic forecasting performance in both short- and long-term contexts. The average MAE, RMSE, and MAPE are calculated for each model for all years. our models achieve more accurate prediction errors compared to traditional online training methods (i.e., Static-STModel). In addition, ST-CRL employs a reinforcement learning model to continuously learn and maintain historical empirical data. It significantly outperforms simple methods such as SVR and GRU and has the performance model of advanced models. Since Static-STModel does not incrementally update the model, it cannot capture new patterns on the traffic network and maintain historical experience, leading to a rapid increase in prediction error. If only new nodes are trained incrementally every year without maintaining the patterns of other nodes, it will lead to a catastrophic forgetting problem. The expressiveness of the model will become unstable. Especially when the patterns of old and new nodes are very different, the prediction results will be poor. ST-CRL can maintain good and stable prediction results after modeling streaming networks, and by prioritizing experience replay and KL analysis, the results show that it can keep learning new traffic patterns while maintaining existing knowledge.

Figures 3(d) and Figures 4(d) show the running time of each model, including the total training time and the training time per epoch, averaging 7 years. It demonstrates that ST-CRL can achieve accurate predictions while maintaining high efficiency, making it possible to use it for display applications. Note that STModel uses the model trained in the first year to predict annual traffic flows thereafter, so it has a training time of 0 after 2011.

\section{Related Work}
\subsection{Traffic Flow Forecasting}
Due to the rapid development of intelligent transportation systems (ITS), predicting the traffic flow has attracted great attention from both industry and academics. Traditional statistical methods using various machine learning techniques, such as VAR, ARIMA and SVR \cite{zivot2006vector,van1996combining,castro2009online}, which can estimate the time-series in the traffic data.But these methods failed to pay attention to the spatial-temporal correlation between different locations. Also it will increase wasted time and storage when training each location with a model. Inspired by the advances in deep learning techniques, researchers have combine deeper networks to overcome the problem of the spatial-temporal data, such as ResNet \cite{he2016deep} and LSTM \cite{S1997lstm} for the traffic prediction. RSTN combines CNN and ConvLSTM \cite{liu2017short} modules through residual connections to capture spatial-temporal and extraneous dependencies. MGSTC \cite{Liu2018MGSTC} explores multiple spatial-temporal correlations through multiple gated spatial-temporal CNN branches, and dynamically combines spatial-temporal features. But these methods are not use for long streaming network. Also there have been some researches on the use of reinforcement learning in the traffic, but the problems are not traffic flow predictions. 
\subsection{Continual Reinforcement Learning}
Continual learning is constant development of increasingly complex behaviors, which is a technique to train the model with the stream data. Existing methods can be divided into three categories: Replay methods \cite{rolnick2019experience}, Regularization-based methods \cite{kirkpatrick2018reply} and Parameter isolation methods.


The main goal of continuous learning is to solve the catastrophic forgetting problem of model training. When conventional deep learning models are trained on new tasks, their performance on old tasks will be significantly reduced, or even devastating. To overcome catastrophic forgetting, a conflict architecture that needs to be resolved, the stability-plasticity dilemma, is proposed. The trained model must have a certain ability to integrate new and old knowledge (stability) on the one hand, and at the same time need to prevent new knowledge from interfering with previous experience (plasticity).

\section{Conclusions}
We study the problem of long-term traffic flow prediction. Due to the contin uous expansion and evolution of the long-term traffic network topology and traffic flow, we use reinforcement learning for modeling and continuous learning for historical data analysis to gradually learn to generate accurate traffic Flow prediction, specifically, we describe the state as a combination of sensor patterns and a transportation network, where interactions are modeled, bidirectional updates based on the practice part, the strategy aims to simulate sensors through the Dueling DQN involved and an improved sampling. Proxy mode to generate accurate traffic flow forecasts. From experiments, we can observe that this strategy better understands sensor patterns and preferences. In addition, the proposed prioritized experience replay strategy and KL analyze historical knowledge and integrate it into continuous reinforcement learning to complete the task of traffic flow prediction. It has been proved to be very effective by experiments. Compared with traditional training methods, it achieves higher prediction accuracy while greatly reducing the training complexity.

\section*{Acknowledgments}

This work is supported by the Natural Science Research Foundation of Jilin Province of China under Grant No. YDZJ202201ZYTS423, the Fundamental Research Funds for the Central Universities (under Grant Nos. 2412022QD040, 93K172022K10, and 2412019ZD013), NSFC (under Grant Nos. 61976050 and 61972384).

\bibliographystyle{IEEEtran}
\bibliography{ref}

\begin{thebibliography}{10}
\providecommand{\url}[1]{#1}
\csname url@samestyle\endcsname
\providecommand{\newblock}{\relax}
\providecommand{\bibinfo}[2]{#2}
\providecommand{\BIBentrySTDinterwordspacing}{\spaceskip=0pt\relax}
\providecommand{\BIBentryALTinterwordstretchfactor}{4}
\providecommand{\BIBentryALTinterwordspacing}{\spaceskip=\fontdimen2\font plus
\BIBentryALTinterwordstretchfactor\fontdimen3\font minus
  \fontdimen4\font\relax}
\providecommand{\BIBforeignlanguage}[2]{{%
\expandafter\ifx\csname l@#1\endcsname\relax
\typeout{** WARNING: IEEEtran.bst: No hyphenation pattern has been}%
\typeout{** loaded for the language `#1'. Using the pattern for}%
\typeout{** the default language instead.}%
\else
\language=\csname l@#1\endcsname
\fi
#2}}
\providecommand{\BIBdecl}{\relax}
\BIBdecl

\bibitem{zhang2017deep}
J.~Zhang, Y.~Zheng, and D.~Qi, ``Deep spatio-temporal residual networks for
  citywide crowd flows prediction,'' in \emph{Thirty-first AAAI conference on
  artificial intelligence}, 2017.

\bibitem{li2017diffusion}
Y.~Li, R.~Yu, C.~Shahabi, and Y.~Liu, ``Diffusion convolutional recurrent
  neural network: Data-driven traffic forecasting,'' \emph{arXiv preprint
  arXiv:1707.01926}, 2017.

\bibitem{wang2022lifelong}
C.~Wang, Y.~Qiu, D.~Gao, and S.~Scherer, ``Lifelong graph learning,'' in
  \emph{Proceedings of the IEEE/CVF Conference on Computer Vision and Pattern
  Recognition}, 2022, pp. 13\,719--13\,728.

\bibitem{yu2017spatio}
B.~Yu, H.~Yin, and Z.~Zhu, ``Spatio-temporal graph convolutional networks: A
  deep learning framework for traffic forecasting,'' \emph{arXiv preprint
  arXiv:1709.04875}, 2017.

\bibitem{zivot2006vector}
E.~Zivot and J.~Wang, ``Vector autoregressive models for multivariate time
  series,'' \emph{Modeling financial time series with S-PLUS{\textregistered}},
  pp. 385--429, 2006.

\bibitem{castro2009online}
M.~Castro-Neto, Y.-S. Jeong, M.-K. Jeong, and L.~D. Han, ``Online-svr for
  short-term traffic flow prediction under typical and atypical traffic
  conditions,'' \emph{Expert systems with applications}, vol.~36, no.~3, pp.
  6164--6173, 2009.

\bibitem{2020Incremental}
P.~Wang, K.~Liu, L.~Jiang, X.~Li, and Y.~Fu, ``Incremental mobile user
  profiling: Reinforcement learning with spatial knowledge graph for modeling
  event streams,'' in \emph{KDD}, 2020.

\bibitem{DBLP:conf/kdd/WangFZWZA18}
P.~Wang, Y.~Fu, J.~Zhang, P.~Wang, Y.~Zheng, and C.~C. Aggarwal, ``You are how
  you drive: Peer and temporal-aware representation learning for driving
  behavior analysis,'' in \emph{SIGKDD}, 2018, pp. 2457--2466.

\bibitem{DBLP:journals/tist/WangFZLL18}
P.~Wang, Y.~Fu, J.~Zhang, X.~Li, and D.~Lin, ``Learning urban community
  structures: {A} collective embedding perspective with periodic
  spatial-temporal mobility graphs,'' \emph{{ACM} Trans. Intell. Syst.
  Technol.}, vol.~9, no.~6, pp. 63:1--63:28, 2018.

\bibitem{DBLP:conf/kdd/WangFXL19}
P.~Wang, Y.~Fu, H.~Xiong, and X.~Li, ``Adversarial substructured representation
  learning for mobile user profiling,'' in \emph{SIGKDD}.\hskip 1em plus 0.5em
  minus 0.4em\relax {ACM}, 2019, pp. 130--138.

\bibitem{wang2020sccwalk}
Y.~Wang, S.~Cai, J.~Chen, and M.~Yin, ``Sccwalk: An efficient local search
  algorithm and its improvements for maximum weight clique problem,''
  \emph{Artificial Intelligence}, vol. 280, p. 103230, 2020.

\bibitem{ShiweiPAN:0}
Y.~W. Z. Z. J. J. M. Y. S.~H. Shiwei~PAN, Yiming~MA, ``An improved
  master-apprentice evolutionary algorithm for minimum independent dominating
  set problem,'' \emph{Frontiers of Computer Science}, pp. 1--17, 2022.

\bibitem{kirkpatrick2018reply}
J.~Kirkpatrick, R.~Pascanu, N.~Rabinowitz, J.~Veness, G.~Desjardins, A.~A.
  Rusu, K.~Milan, J.~Quan, T.~Ramalho, A.~Grabska-Barwinska \emph{et~al.},
  ``Reply to husz{\'a}r: The elastic weight consolidation penalty is
  empirically valid,'' \emph{Proceedings of the National Academy of Sciences},
  vol. 115, no.~11, pp. E2498--E2498, 2018.

\bibitem{scheller2020sample}
C.~Scheller, Y.~Schraner, and M.~Vogel, ``Sample efficient reinforcement
  learning through learning from demonstrations in minecraft,'' in
  \emph{NeurIPS 2019 Competition and Demonstration Track}.\hskip 1em plus 0.5em
  minus 0.4em\relax PMLR, 2020, pp. 67--76.

\bibitem{rusu2016progressive}
A.~A. Rusu, N.~C. Rabinowitz, G.~Desjardins, H.~Soyer, J.~Kirkpatrick,
  K.~Kavukcuoglu, R.~Pascanu, and R.~Hadsell, ``Progressive neural networks,''
  \emph{arXiv preprint arXiv:1606.04671}, 2016.

\bibitem{mnih2013playing}
V.~Mnih, K.~Kavukcuoglu, D.~Silver, A.~Graves, I.~Antonoglou, D.~Wierstra, and
  M.~Riedmiller, ``Playing atari with deep reinforcement learning,''
  \emph{arXiv preprint arXiv:1312.5602}, 2013.

\bibitem{chen2001freeway}
C.~Chen, K.~Petty, A.~Skabardonis, P.~Varaiya, and Z.~Jia, ``Freeway
  performance measurement system: mining loop detector data,''
  \emph{Transportation Research Record}, vol. 1748, no.~1, pp. 96--102, 2001.

\bibitem{cho2014learning}
K.~Cho, B.~Van~Merri{\"e}nboer, C.~Gulcehre, D.~Bahdanau, F.~Bougares,
  H.~Schwenk, and Y.~Bengio, ``Learning phrase representations using rnn
  encoder-decoder for statistical machine translation,'' \emph{arXiv preprint
  arXiv:1406.1078}, 2014.

\bibitem{chen2021trafficstream}
X.~Chen, J.~Wang, and K.~Xie, ``Trafficstream: A streaming traffic flow
  forecasting framework based on graph neural networks and continual
  learning,'' \emph{arXiv preprint arXiv:2106.06273}, 2021.

\bibitem{Wang2006VAR}
E.~Zivot and J.~Wang, ``Vector autoregressive models for multivariate time
  series.'' \emph{Modeling Financial Time Series with S-Plus®, pages
  385–429}, 2006.

\bibitem{M.VanDerVoort1996ARIMA}
M.~D. M.~Van Der~Voort and S.~Watson, ``Combining kohonen maps with arima time
  series models to forecast traffic flow.'' \emph{Transportation Research Part
  C: Emerging Technologies, vol. 4, no. 5, pp.307–318}, 1996.

\bibitem{M.CastroNeto2009SVR}
M.-K.~J. M.~Castro-Neto, Y.-S.~Jeong and L.~D. Han, ``Online-svr for short-term
  traffic flow prediction under typical and atypical traffic conditions.''
  \emph{Expert systems with applications, vol. 36, no. 3, pp. 61646173, 2009},
  2009.

\bibitem{Zhang2016resnet}
S.~R. K.~He, X.~Zhang and J.~Sun, ``Deep residual learning for image
  recognition.'' \emph{in International conference on computer vision and
  pattern recognition (CVPR). IEEE, 2016, pp. 770–778}, 2016.

\bibitem{S1997lstm}
S.~Hochreiter and J.~Schmidhuber, ``Long short-term memory.'' \emph{Neural
  Computation, vol. 9, no. 8, pp. 1735–1780}, 1997.

\bibitem{Liu2017ConvLSTM}
X.~F. Z.~C. Y.~Liu, H.~Zheng, ``Short-term traffic flow prediction with
  conv-lstm, in: 2017 9th international conference on wireless communications
  and signal processing.'' \emph{IEEE, 2017, pp. 1–6}, 2017.

\bibitem{Liu2018MGSTC}
Q.~Y. D.~Chai, L.~Wang, ``Bike flow prediction with multi-graph convolutional
  networks.'' \emph{Proceedings of the 26th ACM SIGSPATIAL International
  Conference on Advances in Geographic Information Systems, 2018, pp.
  397–400}, 2018.

\bibitem{Rolnick2019ExperienceReplay}
J.~S. T. P. .~L. David~Rolnick, Arun~Ahuja and G.~Wayne, ``Experience replay
  for continual learning.'' \emph{arXiv:1811.11682v2}, 2019.

\end{thebibliography}

\end{document}